\documentclass[11pt]{article}
\usepackage{acl2018}
\usepackage{times}
\usepackage{url}
\usepackage{latexsym}
\usepackage{lipsum}
\usepackage{mathtools}
\usepackage{amssymb}
\usepackage{array}
\usepackage[font={footnotesize}]{caption}
\usepackage{xcolor}
\usepackage{multirow}
\usepackage{csquotes}
\usepackage{soul}

\definecolor{8red}{HTML}{A70000}
\definecolor{6red}{HTML}{FF0000}
\definecolor{4red}{HTML}{FF5252}
\definecolor{2red}{HTML}{FF7B7B}
\definecolor{1red}{HTML}{FFBABA}

\DeclareRobustCommand{\hlsix}[1]{{\sethlcolor{6red}\hl{#1}}}
\DeclareRobustCommand{\hlfour}[1]{{\sethlcolor{4red}\hl{#1}}}
\DeclareRobustCommand{\hltwo}[1]{{\sethlcolor{2red}\hl{#1}}}
\DeclareRobustCommand{\hlone}[1]{{\sethlcolor{1red}\hl{#1}}}
\aclfinalcopy

\title{Reasoning with Sarcasm by Reading In-between}

\author{
    Yi Tay$^\dagger$, Luu Anh Tuan$^\psi$, Siu Cheung Hui$^\phi$, Jian Su$^\delta$ \\
    {\tt $^\dagger$ytay017@e.ntu.edu.sg} \\
    {\tt $^\psi$at.luu@i2r.a-star.edu.sg} \\
    {\tt $^\phi$asschui@ntu.edu.sg} \\
    {\tt $^\delta$sujian@i2r.a-star.edu.sg} \\
    $^{\dagger,\phi}$School of Computer Science and Engineering, Nanyang Technological University\\
    $^{\psi, \delta}$A*Star, Institute for Infocomm Research, Singapore
}

\date{}

\begin{document}
\maketitle
\begin{abstract}
Sarcasm is a sophisticated speech act which commonly manifests on social communities such as Twitter and Reddit. The prevalence of sarcasm on the social web is highly disruptive to opinion mining systems due to not only its tendency of polarity flipping but also usage of figurative language. Sarcasm commonly manifests with a contrastive theme either between positive-negative sentiments or between literal-figurative scenarios. In this paper, we revisit the notion of modeling \textit{contrast} in order to reason with sarcasm. More specifically, we propose an attention-based neural model that looks \textit{in-between} instead of \textit{across}, enabling it to explicitly model contrast and incongruity. We conduct extensive experiments on six benchmark datasets from Twitter, Reddit and the Internet Argument Corpus. Our proposed model not only achieves state-of-the-art performance on all datasets but also enjoys improved interpretability.
 \end{abstract}

 \section{Introduction}

Sarcasm, commonly defined as \textit{`An ironical taunt used to express contempt'}, is a challenging NLP problem due to its
highly figurative nature. The usage of sarcasm on the social web is prevalent and can be frequently observed in reviews, microblogs (\textit{tweets}) and online forums. As such, the battle against sarcasm is also regularly cited as one of the key challenges in sentiment analysis and opinion mining applications \cite{pang2008opinion}. Hence, it is both imperative and intuitive that effective sarcasm detectors can bring about numerous benefits to opinion mining applications.

Sarcasm is often associated to several linguistic phenomena such as (1) an explicit contrast between sentiments  or (2) disparity between the conveyed emotion and the author's situation (context). Prior work has considered sarcasm to be a contrast between a positive and negative sentiment \cite{DBLP:conf/emnlp/RiloffQSSGH13}. Consider the following examples:

\begin{enumerate}
\item I absolutely \textit{love} to be \textit{ignored}!
\item Yay!!! The best thing to wake up to is my neighbor's drilling.
\item Perfect movie for people who can't fall asleep.
 \end{enumerate}

Given the examples, we make a crucial observation - Sarcasm relies a lot on the semantic relationships (and contrast) between individual words and phrases in a sentence. For instance, the relationships between phrases $\{$\textit{love}, \textit{ignored}$\}$, $\{$\textit{best}, \textit{drilling}$\}$ and $\{$\textit{movie}, \textit{asleep}$\}$ (in the examples above) richly characterize the nature of sarcasm conveyed, i.e., word pairs tend to be contradictory and more often than not, express a juxtaposition of positive and negative terms. This concept is also explored in \cite{joshi2015harnessing} in which the authors refer to this phenomena as \textit{`incongruity'}. Hence, it would be useful to capture the relationships between selected word pairs in a sentence, i.e., \textit{looking in-between}.

State-of-the-art sarcasm detection systems mainly rely on deep and \textit{sequential} neural networks \cite{DBLP:conf/wassa/GhoshV16,DBLP:conf/coling/ZhangZF16}. In these works, compositional encoders such as gated recurrent units (GRU) \cite{cho2014learning} or long short-term memory (LSTM) \cite{hochreiter1997long} are often employed, with the input document being parsed one word at a time. This has several shortcomings for the sarcasm detection task. Firstly, there is no explicit interaction between word pairs, which hampers its ability to explicitly model contrast, incongruity or juxtaposition of situations. Secondly, it is difficult to capture long-range dependencies. In this case, contrastive situations (or sentiments) which are commonplace in sarcastic language may be hard to detect with simple sequential models.

To overcome the weaknesses of standard sequential models such as recurrent neural networks, our work is based on the intuition that modeling intra-sentence relationships can not only improve classification performance but also pave the way for more explainable neural sarcasm detection methods. In other words, our key intuition manifests itself in the form of an attention-based neural network. While the key idea of most neural attention mechanisms is to focus on relevant words and sub-phrases, it merely looks \textit{across} and does not explicitly capture word-word relationships. Hence, it suffers from the same shortcomings as sequential models.

In this paper, our aim is to combine the effectiveness of state-of-the-art recurrent models while harnessing the intuition of \textit{looking in-between}. We propose a multi-dimensional intra-attention recurrent network that models intricate similarities between each word pair in the sentence. In other words, our novel deep learning model aims to capture \textit{`contrast'} \cite{DBLP:conf/emnlp/RiloffQSSGH13} and \textit{`incongruity'} \cite{joshi2015harnessing} within end-to-end neural networks. Our model can be thought of self-targeted co-attention \cite{DBLP:journals/corr/XiongZS16}, which allows our model to not only capture word-word relationships but also long-range dependencies. Finally, we show that our model produces interpretable attention maps which aid in the explainability of model outputs. To the best of our knowledge, our model is the first attention model that can produce explainable results in the sarcasm detection task.

Briefly, the prime contributions of this work can be summarized as follows:
\begin{itemize}
\item We propose a new state-of-the-art method for sarcasm detection. Our proposed model, the Multi-dimensional Intra-Attention Recurrent Network (MIARN) is strongly based on the intuition of compositional learning by leveraging intra-sentence relationships. To the best of our knowledge, none of the existing state-of-the-art models considered exploiting intra-sentence relationships, solely relying on sequential composition.
\item We conduct extensive experiments on multiple benchmarks from Twitter, Reddit and the Internet Argument Corpus. Our proposed MIARN achieves highly competitive performance on all benchmarks, outperforming existing state-of-the-art models such as GRNN \cite{DBLP:conf/coling/ZhangZF16} and CNN-LSTM-DNN \cite{DBLP:conf/wassa/GhoshV16}.
\end{itemize}

\section{Related Work}
Sarcasm is a complex linguistic phenomena that have long fascinated both linguists and NLP researchers. After all, a better computational understanding of this complicated speech act could potentially bring about numerous benefits to existing opinion mining applications. Across the rich history of research on sarcasm, several theories such as the Situational Disparity Theory \cite{wilson2006pragmatics} and the Negation Theory \cite{giora1995irony} have emerged. In these theories, a common theme is a motif that is strongly grounded in contrast, whether in sentiment, intention, situation or context. \cite{DBLP:conf/emnlp/RiloffQSSGH13} propagates this premise forward, presenting an algorithm strongly based on the intuition that sarcasm arises from a juxtaposition of positive and negative situations.
\subsection{Sarcasm Detection}
Naturally, many works in this area have treated the sarcasm detection task as a standard text classification problem. An extremely comprehensive overview can be found at \cite{joshi2017automatic}. Feature engineering approaches were highly popular, exploiting a wide diverse range of features such as syntactic patterns \cite{tsur2010icwsm}, sentiment lexicons \cite{gonzalez2011identifying}, n-gram \cite{reyes2013multidimensional}, word frequency \cite{barbieri2014modelling}, word shape and pointedness features \cite{ptavcek2014sarcasm}, readability and flips \cite{rajadesingan2015sarcasm}, etc.
Notably, there have been quite a reasonable number of works that propose features based on similarity and contrast. \cite{hernandez2015applying} measured the Wordnet based semantic similarity between words. \cite{joshi2015harnessing} proposed a framework based on explicit and implicit incongruity, utilizing features based on positive-negative patterns. \cite{joshi2016word} proposed similarity features based on word embeddings.
\subsection{Deep Learning for Sarcasm Detection}
Deep learning based methods have recently garnered considerable interest in many areas of NLP research. In our problem domain, \cite{DBLP:conf/coling/ZhangZF16} proposed a recurrent-based model with a gated pooling mechanism for sarcasm detection on Twitter. \cite{DBLP:conf/wassa/GhoshV16} proposed a convolutional long-short-term memory network (CNN-LSTM-DNN) that achieves state-of-the-art performance.

While our work focuses on document-only sarcasm detection, several notable works have proposed models that exploit personality information \cite{DBLP:conf/emnlp/GhoshV17} and user context \cite{amir2016modelling}. Novel methods for sarcasm detection such as gaze / cognitive features \cite{DBLP:conf/acl/MishraKNDB16,DBLP:conf/acl/MishraDB17} have also been explored. \cite{DBLP:conf/acl/PeledR17} proposed a novel framework based on neural machine translation to convert a sequence from sarcastic to non-sarcastic. \cite{felbo2017using} proposed a layer-wise training scheme that utilizes emoji-based distant supervision for sentiment analysis and sarcasm detection tasks.

\subsection{Attention Models for NLP}
In the context of NLP, the key idea of neural attention is to soft select a sequence of words based on their relative importance to the task at hand. Early innovations in attentional paradigms mainly involve neural machine translation \cite{luong2015effective,bahdanau2014neural} for aligning sequence pairs. Attention is also commonplace in many NLP applications such as sentiment classification \cite{chen2016neural,yang2016hierarchical}, aspect-level sentiment analysis \cite{1712.05403,DBLP:conf/cikm/TayTH17,chen2017recurrent} and entailment classification \cite{rocktaschel2015reasoning}. Co-attention / Bi-Attention \cite{DBLP:journals/corr/XiongZS16,seo2016bidirectional} is a form of pairwise attention mechanism that was proposed to model query-document pairs. Intra-attention can be interpreted as a self-targetted co-attention and is seeing a lot promising results in many recent works \cite{vaswani2017attention,DBLP:conf/emnlp/ParikhT0U16,tay2017compare,shen2017disan}. The key idea is to model a sequence against itself, learning to attend while capturing long term dependencies and word-word level interactions. To the best of our knowledge, our work is not only the first work that only applies intra-attention to sarcasm detection but also the first attention model for sarcasm detection.

\section{Our Proposed Approach}
\begin{figure}[ht]
  \centering
    \includegraphics[width=1.0\linewidth]{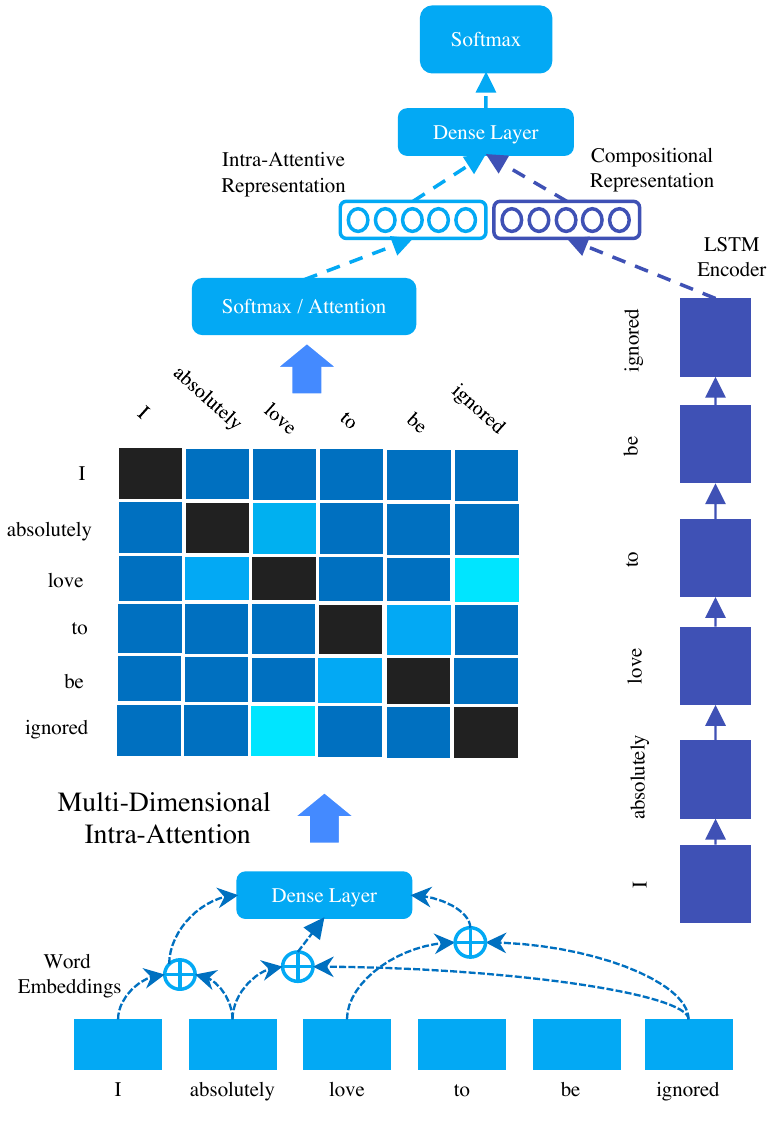}
    \caption{High level overview of our proposed MIARN architecture. MIARN learns two representations, one based on intra-sentence relationships (intra-attentive) and another based on sequential composition (LSTM). Both views are used for prediction.}
    \label{fig:high_level}
\end{figure}

In this section, we describe our proposed model. Figure \ref{fig:high_level} illustrates our overall model architecture.

\subsection{Input Encoding Layer}
Our model accepts a sequence of one-hot encoded vectors as an input. Each one-hot encoded vector corresponds to a single word in the vocabulary. In the
input encoding layer, each one-hot vector is converted into a low-dimensional vector representation (word embedding). The word embeddings are parameterized by an embedding layer $\textbf{W} \in \mathbb{R}^{n \times |V|}$. As such, the output of this layer is a sequence of word embeddings, i.e., $\{ w_1, w_2, \cdots w_{\ell}\}$ where $\ell$ is a predefined maximum sequence length.

\subsection{Multi-dimensional Intra-Attention}
In this section, we describe our multi-dimensional intra-attention mechanism for sarcasm detection. We first begin by describing the standard single-dimensional intra-attention. The multi-dimensional adaptation will be introduced later in this section. The key idea behind this layer is to \textit{look in-between}, i.e., modeling the semantics between each word in the input sequence. We first begin by modeling the relationship of each word pair in the input sequence.
A simple way to achieve this is to use a linear\footnote{Early experiments found that adding nonlinearity here may degrade performance.} transformation layer to project the concatenation of each word embedding \textbf{pair} into a scalar score as follows:
\begin{align}
s_{ij} = W_{a}([w_i;w_j]) + b_{a}
\label{affinity1}
\end{align}
where $W_{a} \in \mathbb{R}^{2n \times 1}, b_{a} \in \mathbb{R}$ are the parameters of this layer. $[.;.]$ is the vector concatenation operator and $s_{ij}$ is a scalar representing the affinity score between word pairs $(w_i, w_j)$. We can easily observe that $s$ is a symmetrical matrix of $\ell \times \ell$ dimensions. In order to learn attention vector $a$, we apply a row-wise max-pooling operator on matrix $s$.
\begin{align}
a = {softmax(\max_{row} s)}
\label{softmax}
\end{align}
where $a \in \mathbb{R}^{\ell}$ is a vector representing the learned intra-attention weights. Then, the vector $a$ is employed to learn weighted representation of $\{w_{1}, w_{2} \cdots w_{\ell}\}$ as follows:
\begin{align}
v_a = \sum_{i=1}^{\ell} \: w_{i} a_{i}
\label{sum}
\end{align}
where $v \in \mathbb{R}^{n}$ is the intra-attentive representation of the input sequence. While other choices of pooling operators may be also employed (e.g., mean-pooling over max-pooling), the choice of max-pooling is empirically motivated. Intuitively, this attention layer learns to pay attention based on a word's \textit{largest} contribution to all words in the sequence. Since our objective is to highlight words that might contribute to the contrastive theories of sarcasm, a more discriminative pooling operator is desirable. Notably, we also mask values of $s$ where $i=j$ such that we do not allow the relationship scores of a word with respect to itself to influence the overall attention weights.

 Furthermore, our network can be considered as an \textit{`inner'} adaptation of neural attention, modeling intra-sentence relationships between the raw word representations instead of representations that have been compositionally manipulated. This allows word-to-word similarity to be modeled \textit{`as it is'} and not be influenced by composition. For example, when using the outputs of a compositional encoder (e.g., LSTM), matching words $n$ and $n+1$ might not be meaningful since they would be relatively similar in terms of semantic composition. For relatively short documents (such as tweets), it is also intuitive that attention typically focuses on the last hidden representation.

Intuitively, the relationships between two words is often not straightforward. Words are complex and often hold more than one meanings (or word senses). As such, it might be beneficial to model \textit{multiple views} between two words. This can be modeled by representing the word pair interaction with a vector instead of a scalar. As such, we propose a multi-dimensional adaptation of the intra-attention mechanism. The key idea here is that each word pair is projected down to a low-dimensional vector before we compute the affinity score, which allows it to not only capture one view (one scalar) but also multiple views. A modification to Equation (\ref{affinity1}) constitutes our Multi-Dimensional Intra-Attention variant.
\begin{align}
s_{ij} = W_{p}(ReLU(W_{q}([w_i;w_j]) + b_{q})) + b_{p}
\end{align}
where $W_{q} \in \mathbb{R}^{n \times k}, W_{p} \in \mathbb{R}^{k \times 1}, b_{q} \in \mathbb{R}^{k}, b_{p} \in \mathbb{R}$ are the parameters of this layer. The final intra-attentive representation is then learned with Equation (\ref{softmax}) and Equation (\ref{sum}) which we do not repeat here for the sake of brevity.
\subsection{Long Short-Term Memory Encoder}
While we are able to simply use the learned representation $v$ for prediction, it is clear that $v$ does not encode compositional information and may miss out on important compositional phrases such as \textit{`not happy'}. Clearly, our intra-attention mechanism simply considers a word-by-word interaction and does not model the input document sequentially. As such, it is beneficial to use a separate compositional encoder for this purpose, i.e., learning compositional representations. To this end, we employ the standard Long Short-Term Memory (LSTM) encoder. The output of an LSTM encoder at each time-step can be briefly defined as:
\begin{align}
h_i &= \text{LSTM}(w, i), \:\:\:\: \forall i \in [1, \dots \ell]
\end{align}
where $\ell$ represents the maximum length of the sequence and $h_i \in \mathbb{R}^{d}$ is the hidden output of the LSTM encoder at time-step $i$. $d$ is the size of the hidden units of the LSTM encoder. LSTM encoders are parameterized by gating mechanisms learned via nonlinear transformations. Since LSTMs are commonplace in standard NLP applications, we omit the technical details for the sake of brevity. Finally, to obtain a compositional representation of the input document, we use $v_{c} = h_{\ell}$ which is the last hidden output of the LSTM encoder. Note that the inputs to the LSTM encoder are the word embeddings right after the input encoding layer and not the output of the intra-attention layer. We found that applying an LSTM on the intra-attentively scaled representations do not yield any benefits.

\subsection{Prediction Layer}
The inputs to the final prediction layer are two representations, namely (1) the intra-attentive representation ($v_a \in \mathbb{R}^{n}$) and (2) the compositional representation ($v_c \in \mathbb{R}^{d}$). This layer learns a joint representation of these two views using a nonlinear projection layer.
\begin{align}
v = ReLU(W_{z}([v_a;v_c]) + b_{z})
\end{align}
where $W_{z} \in \mathbb{R}^{(d+n) \times d}$ and $b_{z} \in \mathbb{R}^{d}$. Finally, we pass $v$ into a Softmax classification layer.
\begin{align}
\hat{y} = Softmax(W_{f}\:v + b_{f})
\end{align}
where $W_{f} \in \mathbb{R}^{d \times 2}, b_{f} \in \mathbb{R}^{2}$ are the parameters of this layer. $\hat{y} \in \mathbb{R}^{2}$ is the output layer of our proposed model.

\subsection{Optimization and Learning}
Our network is trained end-to-end, optimizing the standard binary cross-entropy loss function.

\footnotesize
\begin{align}
J= -\sum^{N}_{i=1} \: [ y_i \log \hat{y}_i + (1-y_i)\log(1-\hat{y}_i)] + R
\end{align}
\normalsize
where $J$ is the cost function, $\hat{y}$ is the output of the network, $R=||\theta||_{L2}$ is the L2 regularization and $\lambda$ is the weight of the regularizer.

\section{Empirical Evaluation}
In this section, we describe our experimental setup and results. Our experiments were designed to answer the following research questions (\textbf{RQ}s).
\begin{itemize}
\item \textbf{RQ1} - Does our proposed approach outperform existing state-of-the-art models?
\item \textbf{RQ2} - What are the impacts of some of the architectural choices of our model? How much does intra-attention contribute to the model performance? Is the Multi-Dimensional adaptation better than the Single-Dimensional adaptation?
\item \textbf{RQ3} - What can we interpret from the intra-attention layers? Does this align with our hypothesis about \textit{looking in-between} and modeling contrast?
\end{itemize}
\subsection{Datasets}
We conduct our experiments on six publicly available benchmark datasets which span across three well-known sources.
\begin{itemize}

\item \textbf{Tweets} - Twitter\footnote{\url{https://twitter.com}}  is a microblogging platform which allows users to post statuses of less than 140 characters. We use two collections for sarcasm detection on tweets. More specifically, we use the dataset obtained from (1) \cite{ptavcek2014sarcasm} in which tweets are trained via hashtag based semi-supervised learning, i.e., hashtags such as \#not, \#sarcasm and \#irony are marked as sarcastic tweets and (2) \cite{DBLP:conf/emnlp/RiloffQSSGH13} in which Tweets are hand annotated and manually checked for sarcasm. For both datasets, we retrieve. Tweets using the Twitter API using the provided tweet IDs.
\item \textbf{Reddit} - Reddit\footnote{\url{https://reddit.com}} is a highly popular social forum and community. Similar to Tweets, sarcastic posts are obtained via the tag `/s' which are marked by the authors themselves. We use two Reddit datasets which are obtained from the subreddits \textit{/r/movies} and \textit{/r/technology} respectively. Datasets are subsets from \cite{khodak2017large}.
\item \textbf{Debates} - We use two datasets\footnote{\url{https://nlds.soe.ucsc.edu/sarcasm1}} from the Internet Argument Corpus (IAC) \cite{lukin2017really} which have been hand annotated for sarcasm. This dataset, unlike the first two, is mainly concerned with long text and provides a diverse comparison from the other datasets. The IAC corpus was designed for research on political debates on online forums. We use the V1 and V2 versions of the sarcasm corpus which are denoted as IAC-V1 and IAC-V2 respectively.

\end{itemize}
The statistics of the datasets used in our experiments is reported in Table \ref{tab:dataset}.
\begin{table}[htbp]
  \centering
  \small

    \begin{tabular}{ccccc}
    \hline
    Dataset & Train & Dev   & Test  & Avg $\ell$ \\
    \hline
    Tweets (\citeauthor{ptavcek2014sarcasm}) & 44017 & 5521  & 5467  & 18 \\
    Tweets (\citeauthor{DBLP:conf/emnlp/RiloffQSSGH13}) & 1369  & 195   & 390   & 14 \\
    Reddit (/r/movies) & 5895  & 655   & 1638  & 12 \\
    Reddit (/r/technology) & 16146 & 1793  & 4571  & 11 \\
    Debates IAC-V1 & 3716  & 464   & 466   & 54 \\
    Debates IAC-V2 & 1549  & 193   & 193   & 64 \\
    \hline
    \end{tabular}%
     \caption{Statistics of datasets used in our experiments.}
  \label{tab:dataset}%
\end{table}%

\begin{table*}[t]
  \centering
\small
    \begin{tabular}{lcccccccc}
    \hline
          & \multicolumn{4}{c}{Tweets \cite{ptavcek2014sarcasm}} & \multicolumn{4}{c}{Tweets \cite{DBLP:conf/emnlp/RiloffQSSGH13}} \\

          \hline
          Model & P     & R     & F1    & Acc   & P     & R     & F1    & Acc \\
          \hline
    NBOW & 80.02 & 79.06 & 79.43 & 80.39 & \textit{71.28} &	62.37	&64.13	&79.23 \\
    Vanilla CNN   & 82.13 & 79.67 & 80.39 & 81.65 & 71.04 & 67.13 & 68.55 & \textit{79.48} \\
    Vanilla LSTM  & \textit{84.62} & 83.21 & 83.67 & \textit{84.50} & 67.33 & 67.20 & 67.27 & 76.27 \\
    Attention LSTM & 84.16 & \textit{85.10} & 83.67 & 84.40 & 68.78 & \textit{68.63} & \textit{68.71} & 77.69 \\
    GRNN (\citeauthor{DBLP:conf/coling/ZhangZF16}) & 84.06 & 83.02 & 83.43 & 84.20 & 66.32 & 64.74 & 65.40 & 76.41 \\
    CNN-LSTM-DNN (\citeauthor{DBLP:conf/wassa/GhoshV16}) & 84.06 & 83.45 & \textit{83.74} & 84.39 & 69.76 & 66.62 & 67.81 & 78.72 \\
    \hline
    SIARN (this paper) & \underline{85.02}	&\underline{84.27}	&\underline{84.59} &	\underline{85.24} &  \textbf{73.82} & \textbf{73.26} & \textbf{73.24} & \textbf{82.31}  \\
    MIARN (this paper) & \textbf{86.13} & \textbf{85.79} & \textbf{86.00} & \textbf{86.47} & \underline{73.34} & 	\underline{68.34}	&\underline{70.10} &	\underline{80.77} \\
    \hline
    \end{tabular}%
      \caption{Experimental Results on Tweets datasets. Best result in is boldface and second best is underlined. Best performing baseline is in \textit{italics}.}
  \label{tab:tweets}%

\vspace{2em}

  \centering

    \begin{tabular}{lcccccccc}
    \hline
  & \multicolumn{4}{c}{Reddit (/r/movies)} & \multicolumn{4}{c}{Reddit (/r/technology)} \\
    \hline
         Model & P & R & F1    & Acc   & P & R & F1    & Acc \\
          \hline
    NBOW & 67.33 & 66.56 & 66.82 & 67.52 & 65.45 & 65.62 & 65.52 & 66.55 \\
    Vanilla CNN   & 65.97 & 65.97 & 65.97 & 66.24 & 65.88 & 62.90 & 62.85 & 66.80 \\
    Vanilla LSTM  & 67.57 & 67.67 & 67.32 & 67.34 & 66.94 & 67.22 & 67.03 & \textit{67.92} \\
    Attention LSTM & 68.11 & 67.87 & 67.94 & 68.37 & \textit{68.20} & \textit{68.78} & \textit{67.44} & 67.22 \\
    GRNN (\citeauthor{DBLP:conf/coling/ZhangZF16}) & 66.16 & 66.16 & 66.16 & 66.42 & 66.56 & 66.73 & 66.66 & 67.65 \\
    CNN-LSTM-DNN (\citeauthor{DBLP:conf/wassa/GhoshV16}) & \textit{68.27} & \textit{67.87} & \textit{67.95} & \textit{68.50} & 66.14 & 66.73 & 65.74 & 66.00 \\
    \hline
    SIARN (this paper) & \underline{69.59} & \textbf{69.48} & \underline{69.52} & \underline{69.84} & \textbf{69.35} & \textbf{70.05} & \textbf{69.22} & \underline{69.57} \\
    MIARN (this paper) & \textbf{69.68} & \underline{69.37} & \textbf{69.54} & \textbf{69.90} & \underline{68.97} & \underline{69.30} & \underline{69.09} & \textbf{69.91} \\
    \hline
    \end{tabular}%
   \caption{Experimental results on Reddit datasets. Best result in is boldface and second best is underlined. Best performing baseline is in \textit{italics}.}
  \label{tab:reddit}%

\vspace{2em}
    \begin{tabular}{lcccccccc}
    \hline
          & \multicolumn{4}{c}{Debates (IAC-V1)}    & \multicolumn{4}{c}{Debates (IAC-V2)} \\
\hline
         Model & P & R & F1    & Acc   & P & R & F1    & Acc \\
          \hline
    NBOW & 57.17 & 57.03 & 57.00    & 57.51 & 66.01 & 66.03 & 66.02 & 66.09 \\
    Vanilla CNN   & 58.21 & 58.00    & 57.95 & 58.55 & 68.45 & 68.18 & 68.21 & 68.56 \\
    Vanilla LSTM  & 54.87 & 54.89 & 54.84 & 54.92 & 68.30  & 63.96 & 60.78 & 62.66 \\
    Attention LSTM & 58.98 & 57.93 & 57.23 & 59.07 & 70.04 & 69.62 & \textit{69.63} & \textit{69.96} \\
    GRNN (\citeauthor{DBLP:conf/coling/ZhangZF16}) & 56.21 & 56.21 & 55.96 & 55.96 & 62.26 & 61.87 & 61.21 & 61.37 \\
    CNN-LSTM-DNN (\citeauthor{DBLP:conf/wassa/GhoshV16})  & 55.50  & 54.60  & 53.31 & 55.96 & 64.31 & 64.33 & 64.31 & 64.38 \\
    \hline
    SIARN (this paper) & \textbf{63.94} & \underline{63.45} & \underline{62.52} & \underline{62.69} & \underline{72.17} & \underline{71.81} &\underline{71.85} & \underline{72.10} \\

    MIARN (this paper) & \underline{63.88} & \textbf{63.71} & \textbf{63.18} & \textbf{63.21} & \textbf{72.92} & \textbf{72.93} & \textbf{72.75} & \textbf{72.75} \\
    \hline
    \end{tabular}%
     \caption{Experimental results on Debates datasets. Best result in is boldface and second best is underlined. Best performing baseline is in \textit{italics}.}
       \label{tab:debates}%
\end{table*}%

\subsection{Compared Methods}
We compare our proposed model with the following algorithms.

\begin{itemize}
\item \textbf{NBOW}  is a simple neural bag-of-words baseline that sums all the word embeddings and passes the summed vector into a simple logistic regression layer.
\item \textbf{CNN} is a vanilla Convolutional Neural Network with max-pooling. CNNs are considered as compositional encoders that capture n-gram features by parameterized sliding windows. The filter width is $3$ and number of filters $f=100$.
\item \textbf{LSTM} is a vanilla Long Short-Term Memory Network. The size of the LSTM cell is set to $d=100$.
\item \textbf{ATT-LSTM (Attention-based LSTM)} is a LSTM model with a neural attention mechanism applied to all the LSTM hidden outputs. We use a similar adaptation to \cite{yang2016hierarchical}, albeit only at the document-level.
\item \textbf{GRNN (Gated Recurrent Neural Network)} is a Bidirectional Gated Recurrent Unit (GRU) model that was proposed for sarcasm detection by \cite{DBLP:conf/coling/ZhangZF16}. GRNN uses a gated pooling mechanism to aggregate the hidden representations from a standard BiGRU model. Since we only compare on document-level sarcasm detection, we do not use the variant of GRNN that exploits user context.
\item \textbf{CNN-LSTM-DNN (Convolutional LSTM + Deep Neural Network)}, proposed by \cite{DBLP:conf/wassa/GhoshV16}, is the state-of-the-art model for sarcasm detection. This model is a combination of a CNN, LSTM and Deep Neural Network via stacking. It stacks two layers of 1D convolution with 2 LSTM layers. The output passes through a deep neural network (DNN) for prediction.
\end{itemize}
Both CNN-LSTM-DNN \cite{DBLP:conf/wassa/GhoshV16} and GRNN \cite{DBLP:conf/coling/ZhangZF16} are state-of-the-art models for document-level sarcasm detection and have outperformed numerous neural and non-neural baselines. In particular, both works have well surpassed feature-based models (Support Vector Machines, etc.), as such we omit comparisons for the sake of brevity and focus comparisons with recent neural models instead. Moreover, since our work focuses only on document-level sarcasm detection, we do not compare against models that use external information such as user profiles, context, personality information \cite{DBLP:conf/emnlp/GhoshV17} or emoji-based distant supervision \cite{felbo2017using}.

For our model, we report results on both multi-dimensional and single-dimensional intra-attention. The two models are named as MIARN and SIARN respectively.
\subsection{Implementation Details and Metrics}
We adopt standard the evaluation metrics for the sarcasm detection task, i.e., macro-averaged F1 and accuracy score. Additionally, we also report precision and recall scores.  All deep learning models are implemented using TensorFlow \cite{tensorflow2015-whitepaper} and optimized on a NVIDIA GTX1070 GPU. Text is preprocessed with NLTK\footnote{\url{https://nltk.org}}'s Tweet tokenizer. Words that only appear once in the entire corpus are removed and marked with the UNK token. Document lengths are truncated at $40,20,80$ tokens for Twitter, Reddit and Debates dataset respectively. Mentions of other users on the Twitter dataset are replaced by `@USER'. Documents with URLs (i.e., containing `http') are removed from the corpus. Documents with less than $5$ tokens are also removed. The learning optimizer used is the RMSProp with an initial learning rate of $0.001$. The L2 regularization is set to $10^{-8}$. We initialize the word embedding layer with GloVe \cite{DBLP:conf/emnlp/PenningtonSM14}. We use the GloVe model trained on 2B Tweets for the Tweets and Reddit dataset. The Glove model trained on Common Crawl is used for the Debates corpus. The size of the word embeddings is fixed at $d=100$ and are fine-tuned during training. In all experiments, we use a development set to select the best hyperparameters. Each model is trained for a total of $30$ epochs and the model is saved each time the performance on the development set is topped. The batch size is tuned amongst $\{128, 256, 512\}$ for all datasets. The only exception is the Tweets dataset from \cite{DBLP:conf/emnlp/RiloffQSSGH13}, in which a batch size of $16$ is used in lieu of the much smaller dataset size. For fair comparison, all models have the same hidden representation size and are set to $100$ for both recurrent and convolutional based models (i.e., number of filters). For MIARN, the size of intra-attention hidden representation is tuned amongst $\{4,8,10,20\}$.

\subsection{Experimental Results}

Table \ref{tab:tweets}, Table \ref{tab:reddit} and Table \ref{tab:debates} reports a performance comparison of all benchmarked models on the Tweets, Reddit and Debates datasets respectively. We observe that our proposed SIARN and MIARN models achieve the best results across all six datasets. The relative improvement differs across domain and datasets. On the Tweets dataset from \cite{ptavcek2014sarcasm}, MIARN achieves about $\approx 2\%-2.2\%$ improvement in terms of F1 and accuracy score when compared against the best baseline. On the other Tweets dataset from \cite{DBLP:conf/emnlp/RiloffQSSGH13}, the performance gain of our proposed model is larger, i.e., $3\%-5\%$ improvement on average over most baselines. Our proposed SIARN and MIARN models achieve very competitive performance on the Reddit datasets, with an average of $\approx 2\%$ margin improvement over the best baselines. Notably, the baselines we compare against are extremely competitive state-of-the-art neural network models. This further reinforces the effectiveness of our proposed approach. Additionally, the performance improvement on Debates (long text) is significantly larger than short text (i.e., Twitter and Reddit). For example, MIARN outperforms GRNN and CNN-LSTM-DNN by $\approx 8\%-10\%$ on both IAC-V1 and IAC-V2. At this note, we can safely put \textbf{RQ1} to rest.

Overall, the performance of MIARN is often marginally better than SIARN (with some exceptions, e.g., \textit{Tweets} dataset from \cite{DBLP:conf/emnlp/RiloffQSSGH13}). We believe that this is attributed to the fact that more complex word-word relationships can be learned by using multi-dimensional values instead of single-dimensional scalars. The performance brought by our additional intra-attentive representations can be further observed by comparing against the vanilla LSTM model. Clearly, removing the intra-attention network reverts our model to the standard LSTM. The performance improvements are encouraging, leading to almost $10\%$ improvement in terms of F1 and accuracy. On datasets with short text, the performance improvement is often a modest $\approx 2\%-3\%$ (\textbf{RQ2}). Notably, our proposed models also perform much better on long text, which can be attributed to the intra-attentive representations explicitly modeling long range dependencies. Intuitively, this is problematic for models that only capture sequential dependencies (e.g., word by word).

Finally, the relative performance of competitor methods are as expected. NBOW performs the worse, since it is just a naive bag-of-words model without any compositional or sequential information. On short text, LSTMs are overall better than CNNs. However, this trend is reversed on long text (i.e., Debates) since the LSTM model may be overburdened by overly long sequences. On short text, we also found that attention (or the gated pooling mechanism from GRNN) did not really help make any significant improvements over the vanilla LSTM model and a qualitative explanation to why this is so is deferred to the next section. However, attention helps for long text (such as debates), resulting in Attention LSTMs becoming the strongest baseline on the Debates datasets. However, our proposed intra-attentive model is both effective on short text and long text, outperforming Attention LSTMs consistently on all datasets.

\subsection{In-depth Model Analysis}
In this section, we present an in-depth analysis of our proposed model. More specifically, we not only aim to showcase the interpretability of our model but also explain how representations are formed. More specifically, we test our model (trained on Tweets dataset by \cite{ptavcek2014sarcasm}) on two examples. We extract the attention maps of three models, namely MIARN, Attention LSTM (ATT-LSTM) and applying Attention mechanism directly on the word embeddings without using a LSTM encoder (ATT-RAW). Table \ref{tab:attexample} shows the visualization of the attention maps.

\begin{table}[htbp]
  \centering
\small
    \begin{tabular}{c|c|c}
    \hline
  Label & Model & Sentence\\
    \hline

     \multirow{3}{*}{{True}} & MIARN &  I totally \hlfour{love} being \hlfour{ignored} !! \\
          & ATT-LSTM &  I totally love being ignored \hlsix{!!}  \\
        & 	ATT-RAW &   \hlone{I} \hltwo{totally} \hlfour{love} \hlone{being} \hlone{ignored} \hlfour{!!}  \\

          \hline
         \multirow{3}{*}{{False}} & MIARN &  \hlone{Being} \hltwo{ignored} \hltwo{sucks} \hlone{big} \hltwo{time}  \\
         &  ATT-LSTM &  Being ignored sucks big \hlsix{time}  \\
       & 	ATT-RAW &  \hlone{Being} \hlone{ignored} \hltwo{sucks} \hltwo{big} \hlone{time}  \\

          \hline

    \end{tabular}%
      \caption{Visualization of normalized attention weights on three different attention models (\textit{Best viewed in color}). The intensity denotes the strength of the attention weight on the word.}
  \label{tab:attexample}%
\end{table}%

In the first example (\textit{true} label), we notice that the attention maps of MIARN are focusing on the words \textit{`love'} and \textit{`ignored'}. This is in concert with our intuition about modeling contrast and incongruity. On the other hand, both ATT-LSTM and ATT-RAW learn very different attention maps. As for ATT-LSTM, the attention weight is focused completely on the last representation - the token `\textit{!!}'. Additionally, we also observed that this is true for many examples in the Tweets and Reddit dataset. We believe that this is the reason why standard neural attention does not help as what the attention mechanism is learning is to select the last representation (i.e., vanilla LSTM). Without the LSTM encoder, the attention weights focus on \textit{`love'} but not \textit{`ignored'}. This fails to capture any concept of contrast or incongruity.

Next, we consider the \textit{false} labeled example. This time, the attention maps of MIARN are not as distinct as before. However, they focus on sentiment-bearing words, composing the words \textit{`ignored sucks'} to form the majority of the intra-attentive representation. This time, passing the vector made up of \textit{`ignored sucks'} allows the subsequent layers to recognize that there is no contrasting situation or sentiment. Similarly, ATT-LSTM focuses on the last word \textit{time} which is totally non-interpretable. On the other hand, ATT-RAW focuses on relatively non-meaningful words such as \textit{`big'}.

Overall, we analyzed two cases (positive and negative labels) and found that MIARN produces very explainable attention maps. In general, we found that MIARN is able to identify contrast and incongruity in sentences, allowing our model to better detect sarcasm. This is facilitated by modeling intra-sentence relationships. Notably, the standard vanilla attention is not explainable or interpretable.

\section{Conclusion}
Based on the intuition of intra-sentence similarity (i.e., \textit{looking in-between}), we proposed a new neural network architecture for sarcasm detection. Our network incorporates a multi-dimensional intra-attention component that learns an intra-attentive representation of the sentence, enabling it to detect contrastive sentiment, situations and incongruity. Extensive experiments over six public benchmarks confirm the empirical effectiveness of our proposed model. Our proposed MIARN model outperforms strong state-of-the-art baselines such as GRNN and CNN-LSTM-DNN. Analysis of the intra-attention scores shows that our model learns highly interpretable attention weights, paving the way for more explainable neural sarcasm detection methods.

\bibliographystyle{acl_natbib}
\bibliography{acl2018}

\begin{thebibliography}{}
\expandafter\ifx\csname natexlab\endcsname\relax\def\natexlab#1{#1}\fi

\bibitem[{Abadi et~al.(2015)Abadi, Agarwal, Barham, Brevdo, Chen, Citro,
  Corrado, Davis, Dean, Devin, Ghemawat, Goodfellow, Harp, Irving, Isard, Jia,
  Jozefowicz, Kaiser, Kudlur, Levenberg, Man\'{e}, Monga, Moore, Murray, Olah,
  Schuster, Shlens, Steiner, Sutskever, Talwar, Tucker, Vanhoucke, Vasudevan,
  Vi\'{e}gas, Vinyals, Warden, Wattenberg, Wicke, Yu, and
  Zheng}]{tensorflow2015-whitepaper}
Mart\'{\i}n Abadi, Ashish Agarwal, Paul Barham, Eugene Brevdo, Zhifeng Chen,
  Craig Citro, Greg~S. Corrado, Andy Davis, Jeffrey Dean, Matthieu Devin,
  Sanjay Ghemawat, Ian Goodfellow, Andrew Harp, Geoffrey Irving, Michael Isard,
  Yangqing Jia, Rafal Jozefowicz, Lukasz Kaiser, Manjunath Kudlur, Josh
  Levenberg, Dan Man\'{e}, Rajat Monga, Sherry Moore, Derek Murray, Chris Olah,
  Mike Schuster, Jonathon Shlens, Benoit Steiner, Ilya Sutskever, Kunal Talwar,
  Paul Tucker, Vincent Vanhoucke, Vijay Vasudevan, Fernanda Vi\'{e}gas, Oriol
  Vinyals, Pete Warden, Martin Wattenberg, Martin Wicke, Yuan Yu, and Xiaoqiang
  Zheng. 2015.
\newblock {TensorFlow}: Large-scale machine learning on heterogeneous systems.
\newblock Software available from tensorflow.org.

\bibitem[{Amir et~al.(2016)Amir, Wallace, Lyu, and Silva}]{amir2016modelling}
Silvio Amir, Byron~C Wallace, Hao Lyu, and Paula Carvalho M{\'a}rio~J Silva.
  2016.
\newblock Modelling context with user embeddings for sarcasm detection in
  social media.
\newblock {\em arXiv preprint arXiv:1607.00976\/} .

\bibitem[{Bahdanau et~al.(2014)Bahdanau, Cho, and Bengio}]{bahdanau2014neural}
Dzmitry Bahdanau, Kyunghyun Cho, and Yoshua Bengio. 2014.
\newblock Neural machine translation by jointly learning to align and
  translate.
\newblock {\em arXiv preprint arXiv:1409.0473\/} .

\bibitem[{Barbieri et~al.(2014)Barbieri, Saggion, and
  Ronzano}]{barbieri2014modelling}
Francesco Barbieri, Horacio Saggion, and Francesco Ronzano. 2014.
\newblock Modelling sarcasm in twitter, a novel approach.
\newblock In {\em Proceedings of the 5th Workshop on Computational Approaches
  to Subjectivity, Sentiment and Social Media Analysis\/}. pages 50--58.

\bibitem[{Chen et~al.(2016)Chen, Sun, Tu, Lin, and Liu}]{chen2016neural}
Huimin Chen, Maosong Sun, Cunchao Tu, Yankai Lin, and Zhiyuan Liu. 2016.
\newblock Neural sentiment classification with user and product attention.
\newblock In {\em Proceedings of the 2016 Conference on Empirical Methods in
  Natural Language Processing\/}. pages 1650--1659.

\bibitem[{Chen et~al.(2017)Chen, Sun, Bing, and Yang}]{chen2017recurrent}
Peng Chen, Zhongqian Sun, Lidong Bing, and Wei Yang. 2017.
\newblock Recurrent attention network on memory for aspect sentiment analysis.
\newblock In {\em Proceedings of the 2017 Conference on Empirical Methods in
  Natural Language Processing\/}. pages 452--461.

\bibitem[{Cho et~al.(2014)Cho, Van~Merri{\"e}nboer, Gulcehre, Bahdanau,
  Bougares, Schwenk, and Bengio}]{cho2014learning}
Kyunghyun Cho, Bart Van~Merri{\"e}nboer, Caglar Gulcehre, Dzmitry Bahdanau,
  Fethi Bougares, Holger Schwenk, and Yoshua Bengio. 2014.
\newblock Learning phrase representations using rnn encoder-decoder for
  statistical machine translation.
\newblock {\em arXiv preprint arXiv:1406.1078\/} .

\bibitem[{Felbo et~al.(2017)Felbo, Mislove, S{\o}gaard, Rahwan, and
  Lehmann}]{felbo2017using}
Bjarke Felbo, Alan Mislove, Anders S{\o}gaard, Iyad Rahwan, and Sune Lehmann.
  2017.
\newblock Using millions of emoji occurrences to learn any-domain
  representations for detecting sentiment, emotion and sarcasm.
\newblock {\em arXiv preprint arXiv:1708.00524\/} .

\bibitem[{Ghosh and Veale(2016)}]{DBLP:conf/wassa/GhoshV16}
Aniruddha Ghosh and Tony Veale. 2016.
\newblock \href{http://aclweb.org/anthology/W/W16/W16-0425.pdf}{Fracking
  sarcasm using neural network}.
\newblock In {\em Proceedings of the 7th Workshop on Computational Approaches
  to Subjectivity, Sentiment and Social Media Analysis, WASSA@NAACL-HLT 2016,
  June 16, 2016, San Diego, California, {USA}\/}. pages 161--169.
\newblock
  \href{http://aclweb.org/anthology/W/W16/W16-0425.pdf}{http://aclweb.org/anthology/W/W16/W16-0425.pdf}.

\bibitem[{Ghosh and Veale(2017)}]{DBLP:conf/emnlp/GhoshV17}
Aniruddha Ghosh and Tony Veale. 2017.
\newblock Magnets for sarcasm: Making sarcasm detection timely, contextual and
  very personal.
\newblock In {\em Proceedings of the 2017 Conference on Empirical Methods in
  Natural Language Processing, {EMNLP} 2017, Copenhagen, Denmark, September
  9-11, 2017\/}. pages 482--491.

\bibitem[{Giora(1995)}]{giora1995irony}
Rachel Giora. 1995.
\newblock On irony and negation.
\newblock {\em Discourse processes\/} 19(2):239--264.

\bibitem[{Gonz{\'a}lez-Ib{\'a}nez et~al.(2011)Gonz{\'a}lez-Ib{\'a}nez, Muresan,
  and Wacholder}]{gonzalez2011identifying}
Roberto Gonz{\'a}lez-Ib{\'a}nez, Smaranda Muresan, and Nina Wacholder. 2011.
\newblock Identifying sarcasm in twitter: a closer look.
\newblock In {\em Proceedings of the 49th Annual Meeting of the Association for
  Computational Linguistics: Human Language Technologies: Short Papers-Volume
  2\/}. Association for Computational Linguistics, pages 581--586.

\bibitem[{Hern{\'a}ndez-Far{\'\i}as et~al.(2015)Hern{\'a}ndez-Far{\'\i}as,
  Bened{\'\i}, and Rosso}]{hernandez2015applying}
Iraz{\'u} Hern{\'a}ndez-Far{\'\i}as, Jos{\'e}-Miguel Bened{\'\i}, and Paolo
  Rosso. 2015.
\newblock Applying basic features from sentiment analysis for automatic irony
  detection.
\newblock In {\em Iberian Conference on Pattern Recognition and Image
  Analysis\/}. Springer, pages 337--344.

\bibitem[{Hochreiter and Schmidhuber(1997)}]{hochreiter1997long}
Sepp Hochreiter and J{\"u}rgen Schmidhuber. 1997.
\newblock Long short-term memory.
\newblock {\em Neural computation\/} 9(8):1735--1780.

\bibitem[{Joshi et~al.(2017)Joshi, Bhattacharyya, and
  Carman}]{joshi2017automatic}
Aditya Joshi, Pushpak Bhattacharyya, and Mark~J Carman. 2017.
\newblock Automatic sarcasm detection: A survey.
\newblock {\em ACM Computing Surveys (CSUR)\/} 50(5):73.

\bibitem[{Joshi et~al.(2015)Joshi, Sharma, and
  Bhattacharyya}]{joshi2015harnessing}
Aditya Joshi, Vinita Sharma, and Pushpak Bhattacharyya. 2015.
\newblock Harnessing context incongruity for sarcasm detection.
\newblock In {\em Proceedings of the 53rd Annual Meeting of the Association for
  Computational Linguistics and the 7th International Joint Conference on
  Natural Language Processing (Volume 2: Short Papers)\/}. volume~2, pages
  757--762.

\bibitem[{Joshi et~al.(2016)Joshi, Tripathi, Patel, Bhattacharyya, and
  Carman}]{joshi2016word}
Aditya Joshi, Vaibhav Tripathi, Kevin Patel, Pushpak Bhattacharyya, and Mark
  Carman. 2016.
\newblock Are word embedding-based features useful for sarcasm detection?
\newblock {\em arXiv preprint arXiv:1610.00883\/} .

\bibitem[{Khodak et~al.(2017)Khodak, Saunshi, and Vodrahalli}]{khodak2017large}
Mikhail Khodak, Nikunj Saunshi, and Kiran Vodrahalli. 2017.
\newblock A large self-annotated corpus for sarcasm.
\newblock {\em arXiv preprint arXiv:1704.05579\/} .

\bibitem[{Lukin and Walker(2017)}]{lukin2017really}
Stephanie Lukin and Marilyn Walker. 2017.
\newblock Really? well. apparently bootstrapping improves the performance of
  sarcasm and nastiness classifiers for online dialogue.
\newblock {\em arXiv preprint arXiv:1708.08572\/} .

\bibitem[{Luong et~al.(2015)Luong, Pham, and Manning}]{luong2015effective}
Minh-Thang Luong, Hieu Pham, and Christopher~D Manning. 2015.
\newblock Effective approaches to attention-based neural machine translation.
\newblock {\em arXiv preprint arXiv:1508.04025\/} .

\bibitem[{Mishra et~al.(2017)Mishra, Dey, and
  Bhattacharyya}]{DBLP:conf/acl/MishraDB17}
Abhijit Mishra, Kuntal Dey, and Pushpak Bhattacharyya. 2017.
\newblock \href{https://doi.org/10.18653/v1/P17-1035}{Learning cognitive
  features from gaze data for sentiment and sarcasm classification using
  convolutional neural network}.
\newblock In {\em Proceedings of the 55th Annual Meeting of the Association for
  Computational Linguistics, {ACL} 2017, Vancouver, Canada, July 30 - August 4,
  Volume 1: Long Papers\/}. pages 377--387.
\newblock
  \href{https://doi.org/10.18653/v1/P17-1035}{https://doi.org/10.18653/v1/P17-1035}.

\bibitem[{Mishra et~al.(2016)Mishra, Kanojia, Nagar, Dey, and
  Bhattacharyya}]{DBLP:conf/acl/MishraKNDB16}
Abhijit Mishra, Diptesh Kanojia, Seema Nagar, Kuntal Dey, and Pushpak
  Bhattacharyya. 2016.
\newblock \href{http://aclweb.org/anthology/P/P16/P16-1104.pdf}{Harnessing
  cognitive features for sarcasm detection}.
\newblock In {\em Proceedings of the 54th Annual Meeting of the Association for
  Computational Linguistics, {ACL} 2016, August 7-12, 2016, Berlin, Germany,
  Volume 1: Long Papers\/}.
\newblock
  \href{http://aclweb.org/anthology/P/P16/P16-1104.pdf}{http://aclweb.org/anthology/P/P16/P16-1104.pdf}.

\bibitem[{Pang et~al.(2008)Pang, Lee et~al.}]{pang2008opinion}
Bo~Pang, Lillian Lee, et~al. 2008.
\newblock Opinion mining and sentiment analysis.
\newblock {\em Foundations and Trends{\textregistered} in Information
  Retrieval\/} 2(1--2):1--135.

\bibitem[{Parikh et~al.(2016)Parikh, T{\"{a}}ckstr{\"{o}}m, Das, and
  Uszkoreit}]{DBLP:conf/emnlp/ParikhT0U16}
Ankur~P. Parikh, Oscar T{\"{a}}ckstr{\"{o}}m, Dipanjan Das, and Jakob
  Uszkoreit. 2016.
\newblock A decomposable attention model for natural language inference.
\newblock In {\em Proceedings of the 2016 Conference on Empirical Methods in
  Natural Language Processing, {EMNLP} 2016, Austin, Texas, USA, November 1-4,
  2016\/}. pages 2249--2255.

\bibitem[{Peled and Reichart(2017)}]{DBLP:conf/acl/PeledR17}
Lotem Peled and Roi Reichart. 2017.
\newblock \href{https://doi.org/10.18653/v1/P17-1155}{Sarcasm {SIGN:}
  interpreting sarcasm with sentiment based monolingual machine translation}.
\newblock In {\em Proceedings of the 55th Annual Meeting of the Association for
  Computational Linguistics, {ACL} 2017, Vancouver, Canada, July 30 - August 4,
  Volume 1: Long Papers\/}. pages 1690--1700.
\newblock
  \href{https://doi.org/10.18653/v1/P17-1155}{https://doi.org/10.18653/v1/P17-1155}.

\bibitem[{Pennington et~al.(2014)Pennington, Socher, and
  Manning}]{DBLP:conf/emnlp/PenningtonSM14}
Jeffrey Pennington, Richard Socher, and Christopher~D. Manning. 2014.
\newblock Glove: Global vectors for word representation.
\newblock In {\em Proceedings of the 2014 Conference on Empirical Methods in
  Natural Language Processing, {EMNLP} 2014, October 25-29, 2014, Doha, Qatar,
  {A} meeting of SIGDAT, a Special Interest Group of the {ACL}\/}. pages
  1532--1543.

\bibitem[{Pt{\'a}{\v{c}}ek et~al.(2014)Pt{\'a}{\v{c}}ek, Habernal, and
  Hong}]{ptavcek2014sarcasm}
Tom{\'a}{\v{s}} Pt{\'a}{\v{c}}ek, Ivan Habernal, and Jun Hong. 2014.
\newblock Sarcasm detection on czech and english twitter.
\newblock In {\em Proceedings of COLING 2014, the 25th International Conference
  on Computational Linguistics: Technical Papers\/}. pages 213--223.

\bibitem[{Rajadesingan et~al.(2015)Rajadesingan, Zafarani, and
  Liu}]{rajadesingan2015sarcasm}
Ashwin Rajadesingan, Reza Zafarani, and Huan Liu. 2015.
\newblock Sarcasm detection on twitter: A behavioral modeling approach.
\newblock In {\em Proceedings of the Eighth ACM International Conference on Web
  Search and Data Mining\/}. ACM, pages 97--106.

\bibitem[{Reyes et~al.(2013)Reyes, Rosso, and
  Veale}]{reyes2013multidimensional}
Antonio Reyes, Paolo Rosso, and Tony Veale. 2013.
\newblock A multidimensional approach for detecting irony in twitter.
\newblock {\em Language resources and evaluation\/} 47(1):239--268.

\bibitem[{Riloff et~al.(2013)Riloff, Qadir, Surve, Silva, Gilbert, and
  Huang}]{DBLP:conf/emnlp/RiloffQSSGH13}
Ellen Riloff, Ashequl Qadir, Prafulla Surve, Lalindra~De Silva, Nathan Gilbert,
  and Ruihong Huang. 2013.
\newblock \href{http://aclweb.org/anthology/D/D13/D13-1066.pdf}{Sarcasm as
  contrast between a positive sentiment and negative situation}.
\newblock In {\em Proceedings of the 2013 Conference on Empirical Methods in
  Natural Language Processing, {EMNLP} 2013, 18-21 October 2013, Grand Hyatt
  Seattle, Seattle, Washington, USA, {A} meeting of SIGDAT, a Special Interest
  Group of the {ACL}\/}. pages 704--714.
\newblock
  \href{http://aclweb.org/anthology/D/D13/D13-1066.pdf}{http://aclweb.org/anthology/D/D13/D13-1066.pdf}.

\bibitem[{Rockt{\"a}schel et~al.(2015)Rockt{\"a}schel, Grefenstette, Hermann,
  Ko{\v{c}}isk{\`y}, and Blunsom}]{rocktaschel2015reasoning}
Tim Rockt{\"a}schel, Edward Grefenstette, Karl~Moritz Hermann, Tom{\'a}{\v{s}}
  Ko{\v{c}}isk{\`y}, and Phil Blunsom. 2015.
\newblock Reasoning about entailment with neural attention.
\newblock {\em arXiv preprint arXiv:1509.06664\/} .

\bibitem[{Seo et~al.(2016)Seo, Kembhavi, Farhadi, and
  Hajishirzi}]{seo2016bidirectional}
Minjoon Seo, Aniruddha Kembhavi, Ali Farhadi, and Hannaneh Hajishirzi. 2016.
\newblock Bidirectional attention flow for machine comprehension.
\newblock {\em arXiv preprint arXiv:1611.01603\/} .

\bibitem[{Shen et~al.(2017)Shen, Zhou, Long, Jiang, Pan, and
  Zhang}]{shen2017disan}
Tao Shen, Tianyi Zhou, Guodong Long, Jing Jiang, Shirui Pan, and Chengqi Zhang.
  2017.
\newblock Disan: Directional self-attention network for rnn/cnn-free language
  understanding.
\newblock {\em arXiv preprint arXiv:1709.04696\/} .

\bibitem[{Tay et~al.(2018s)Tay, Luu, and Hui}]{1712.05403}
Yi~Tay, Anh~Tuan Luu, and Siu~Cheung Hui. 2018s.
\newblock Learning to attend via word-aspect associative fusion for
  aspect-based sentiment analysis.
\newblock In {\em In Proceedings of the AAAI 2018, 5956-5963\/}.

\bibitem[{Tay et~al.(2017{\natexlab{a}})Tay, Tuan, and Hui}]{tay2017compare}
Yi~Tay, Luu~Anh Tuan, and Siu~Cheung Hui. 2017{\natexlab{a}}.
\newblock A compare-propagate architecture with alignment factorization for
  natural language inference.
\newblock {\em arXiv preprint arXiv:1801.00102\/} .

\bibitem[{Tay et~al.(2017{\natexlab{b}})Tay, Tuan, and
  Hui}]{DBLP:conf/cikm/TayTH17}
Yi~Tay, Luu~Anh Tuan, and Siu~Cheung Hui. 2017{\natexlab{b}}.
\newblock \href{https://doi.org/10.1145/3132847.3132936}{Dyadic memory networks
  for aspect-based sentiment analysis}.
\newblock In {\em Proceedings of the 2017 {ACM} on Conference on Information
  and Knowledge Management, {CIKM} 2017, Singapore, November 06 - 10, 2017\/}.
  pages 107--116.
\newblock
  \href{https://doi.org/10.1145/3132847.3132936}{https://doi.org/10.1145/3132847.3132936}.

\bibitem[{Tsur et~al.(2010)Tsur, Davidov, and Rappoport}]{tsur2010icwsm}
Oren Tsur, Dmitry Davidov, and Ari Rappoport. 2010.
\newblock Icwsm-a great catchy name: Semi-supervised recognition of sarcastic
  sentences in online product reviews.

\bibitem[{Vaswani et~al.(2017)Vaswani, Shazeer, Parmar, Uszkoreit, Jones,
  Gomez, Kaiser, and Polosukhin}]{vaswani2017attention}
Ashish Vaswani, Noam Shazeer, Niki Parmar, Jakob Uszkoreit, Llion Jones,
  Aidan~N Gomez, {\L}ukasz Kaiser, and Illia Polosukhin. 2017.
\newblock Attention is all you need.
\newblock In {\em Advances in Neural Information Processing Systems\/}. pages
  6000--6010.

\bibitem[{Wilson(2006)}]{wilson2006pragmatics}
Deirdre Wilson. 2006.
\newblock The pragmatics of verbal irony: Echo or pretence?
\newblock {\em Lingua\/} 116(10):1722--1743.

\bibitem[{Xiong et~al.(2016)Xiong, Zhong, and
  Socher}]{DBLP:journals/corr/XiongZS16}
Caiming Xiong, Victor Zhong, and Richard Socher. 2016.
\newblock Dynamic coattention networks for question answering.
\newblock {\em CoRR\/} abs/1611.01604.

\bibitem[{Yang et~al.(2016)Yang, Yang, Dyer, He, Smola, and
  Hovy}]{yang2016hierarchical}
Zichao Yang, Diyi Yang, Chris Dyer, Xiaodong He, Alexander~J Smola, and
  Eduard~H Hovy. 2016.
\newblock Hierarchical attention networks for document classification.

\bibitem[{Zhang et~al.(2016)Zhang, Zhang, and Fu}]{DBLP:conf/coling/ZhangZF16}
Meishan Zhang, Yue Zhang, and Guohong Fu. 2016.
\newblock \href{http://aclweb.org/anthology/C/C16/C16-1231.pdf}{Tweet sarcasm
  detection using deep neural network}.
\newblock In {\em {COLING} 2016, 26th International Conference on Computational
  Linguistics, Proceedings of the Conference: Technical Papers, December 11-16,
  2016, Osaka, Japan\/}. pages 2449--2460.
\newblock
  \href{http://aclweb.org/anthology/C/C16/C16-1231.pdf}{http://aclweb.org/anthology/C/C16/C16-1231.pdf}.

\end{thebibliography}








\end{document}